# The Computational Complexity of Dominance and Consistency in CP-Nets

**Judy Goldsmith**                                                                                    GOLDSMIT@CS.UKY.EDU
*Department of Computer Science*
*University of Kentucky*
*Lexington, KY 40506-0046, USA*

**Jérôme Lang**                                                                                              LANG@IRIT.FR
*IRIT*
*Université de Toulouse, UPS*
*31062 Toulouse Cedex, France*

**Miroslaw Truszczyński**                                                                            MIREK@CS.UKY.EDU
*Department of Computer Science*
*University of Kentucky*
*Lexington, KY 40506-0046, USA*

**Nic Wilson**                                                                                          N.WILSON@4C.UCC.IE
*Cork Constraint Computation Centre*
*University College*
*Cork, Ireland*

## Abstract

We investigate the computational complexity of testing dominance and consistency in CP-nets. Previously, the complexity of dominance has been determined for restricted classes in which the dependency graph of the CP-net is acyclic. However, there are preferences of interest that define cyclic dependency graphs; these are modeled with general CP-nets. In our main results, we show here that both dominance and consistency for general CP-nets are PSPACE-complete. We then consider the concept of strong dominance, dominance equivalence and dominance incomparability, and several notions of optimality, and identify the complexity of the corresponding decision problems. The reductions used in the proofs are from STRIPS planning, and thus reinforce the earlier established connections between both areas.

## 1. Introduction

The problems of eliciting, representing and computing with preferences over a multi-attribute domain arise in many fields such as planning, design, and group decision making. However, in a multi-attribute preference domain, such computations may be nontrivial, as we show here for the CP-net representation. Natural questions that arise in a preference domain are, "Is this item preferred to that one?", and "Is this set of preferences consistent?" More formally, a set of preferences is consistent if and only if no item is preferred to itself. We assume that preferences are *transitive*, i.e., if α is preferred to β, and β is preferred to γ, then α is preferred to γ.

An explicit representation of a preference ordering of elements, also called *outcomes*, of such multi-variable domains is exponentially large in the number of attributes. Therefore, AI researchers have developed languages for representing preference orderings in a succinct way. The formalism of CP-nets (Boutilier, Brafman, Hoos, & Poole, 1999) is among the most popular ones. A CP-net provides a succinct representation of preference ordering on outcomes in terms of local preference statements of the form $p : x_i > x_j$, where $x_i, x_j$ are values of a variable $X$ and $p$ is a logical condition. Informally, a preference statement $p : x_i > x_j$ means that given $p$, $x_i$ is strictly preferred to $x_j$ *ceteris paribus*, that is, *all other things being equal*. The meaning of a CP-net is given by a certain ordering relation, called *dominance*, on the set of outcomes, derived from such reading of preference statements. If one outcome dominates another, we say that the dominant one is *preferred*.

Reasoning about the preference ordering (dominance relation) expressed by a CP-net is far from easy. The key problems include *dominance testing* and *consistency testing*. In the first problem, given a CP-net and two outcomes α and β, we want to decide whether β dominates α. The second problem asks whether there is a dominance cycle in the dominance ordering defined by an input CP-net, that is, whether there is an outcome that dominates (is preferred to) itself.

We study the computational complexity of these two problems. The results obtained prior to this work concerned only restricted classes of CP-nets, all requiring that the graph of variable dependencies implied by preference statements in the CP-net be *acyclic*. Under certain assumptions, the dominance-testing problem is in NP and, under some additional assumptions, even in P (Domshlak & Brafman, 2002; Boutilier, Brafman, Domshlak, Hoos, & Poole, 2004a). We show that the complexity in the general case is PSPACE-complete, and this holds even for the propositional case, by exhibiting in Section 4 a PSPACE-hardness proof for dominance testing.

We then turn to consistency testing. While acyclic CP-nets are guaranteed to be consistent, this is not the case with general CP-nets (Domshlak & Brafman, 2002; Brafman & Dimopoulos, 2004). In Section 5, we show that consistency testing is as hard as dominance testing.

In the following two sections we study decision problems related to dominance and optimality in CP-nets. First, we consider the complexity of deciding strict dominance, dominance equivalence and dominance incomparability of outcomes in a CP-net. Then, we study the complexity of deciding the optimality of outcomes, and the existence of optimal outcomes, for several notions of optimality.

To prove the hardness part of the results, we first establish the PSPACE-hardness of some problems related to propositional STRIPS planning. We then show that these problems can be reduced to CP-net dominance and consistency testing by exploiting connections between actions in STRIPS planning and preference statements in CP-nets.

The complexity results in this paper address CP-nets whose dominance relation may contain cycles. Most earlier work has concentrated on the acyclic model. However, as argued earlier, for instance by Domshlak and Brafman (2002), acyclic CP-nets are not sufficiently expressive to capture human preferences on even some simple domains.[1] Consider, for instance, a diner who has to choose either red or white wine, and either fish or meat. Given red wine, they prefer meat, and conversely, given meat they prefer red wine. On the other hand, given white wine, they prefer fish, and conversely, given fish they prefer white wine. This gives a consistent cyclic CP-net, and there is no acyclic CP-net giving rise to the same preferences on outcomes. So, such cyclicity of preference variables does not necessarily lead to a cyclic order on outcomes.

---

1. We do not mean to say that cyclic CP-nets are sufficient to capture *all* possible human preferences on simple domains – this is obviously not true. However, we note that every preference relation *extends* the preference relation induced by some CP-net with possibly cyclic dependencies. Not only is this property no longer true when cyclic dependencies are precluded but, in the case of binary variables, the number of linear orders that extends some acyclic CP-net is exponentially smaller than the number of all linear orders (Xia, Conitzer, & Lang, 2008).

We assume some familiarity with the complexity class PSPACE. We refer to Papadimitriou (1994) for details. In particular, we later use the identities NPSPACE = PSPACE = coPSPACE.

In several places, we will consider versions of decision problem, in which input instances are assumed to have some additional property. Such problems are usually formulated in the following way: "$Q$, given $R$"[2]. We first note that "$Q$, given $R$" is not the same problem as "$Q$ and $R$". Let us recall the definition of a decision problem as presented by Ausiello et al. (1999). A *decision problem* is a pair $\mathcal{P} = \langle I_\mathcal{P}, Y_\mathcal{P} \rangle$ where $I_\mathcal{P}$ is a set of strings (formally, a subset of $\Sigma^*$, where $\Sigma$ is a finite alphabet), The decision problem $\mathcal{P} = \langle I_\mathcal{P}, Y_\mathcal{P} \rangle$ reads as follows: given a string $x \in I_\mathcal{P}$, decide whether $x \in Y_\mathcal{P}$. A problem $\langle I_\mathcal{P}, Y_\mathcal{P} \rangle$ is in a complexity class C if the language $Y_\mathcal{P} \subseteq \Sigma^*$ is in C (this does not depend on $I_\mathcal{P}$). A problem $\langle I_Q, Y_Q \rangle$ is reducible to $\langle I_\mathcal{P}, Y_\mathcal{P} \rangle$ if there is a polynomial-time function $F$ such that (1) for every $x \in I_Q$, $F(x) \in I_\mathcal{P}$, and (2) for every $x \in I_Q$, $x \in Y_Q$ if and only if $F(x) \in Y_\mathcal{P}$. Thus, if $\mathcal{P}$ is the decision problem "$Q$, given $R$", then $I_\mathcal{P}$ is the set of all strings satisfying $R$, while $Y_\mathcal{P}$ is the set of all strings satisfying $R \cap Q$. For all such problems, it is granted that the input belongs to $R$; to solve them we do not have to check that the input string is indeed an element of $R$. Such problems "$Q$, given $R$" are widespread in the literature. However, in most cases, $R$ is a very simple property, that can be checked in polynomial (and often linear) time, such as "decide whether a graph possesses a Hamiltonian cycle, given that every vertex has a degree at most 3". Here, however, we will consider several problems "$Q$, given $R$" where $R$ itself is not in the class P (unless the polynomial hierarchy collapses). However, as we said above, the complexity of recognizing whether a given string is in $R$ does not matter. In other words, the complexity of "$Q$, given $R$" is the same, whether $R$ can be recognized in unit time or is PSPACE-complete. We will come back to this when the first such problem appears in the paper (cf. the proof of Proposition 5). In no case that we consider is the complexity of $R$ greater than the complexity of $Q$.

A part of this paper (up to Section 5) is an extended version of our earlier conference publication (Goldsmith, Lang, Truszczyński, & Wilson, 2005). Sections 6 and 7 are entirely new.

## 2. Generalized Propositional CP-Nets

Let $V = \{x_1, \ldots, x_n\}$ be a finite set of *variables*. For each variable $x \in V$, we assume a finite *domain* $D_x$ of *values*. An *outcome* is an $n$-tuple $(d_1, \ldots, d_n)$ of $D_{x_1} \times \cdots \times D_{x_n}$.

In this paper, we focus on *propositional* variables: variables with *binary* domains. Let $V$ be a finite set of propositional variables. For every $x \in V$, we set $D_x = \{x, \neg x\}$ (thus, we overload the notation and write $x$ both for the variable and for one of its values). We refer to $x$ and $\neg x$ as literals. Given a literal $l$ we write $\neg l$ to denote the dual literal to $l$. The focus on binary variables makes the presentation clearer and has no impact on our complexity results.

We also note that in the case of binary domains, we often identify an outcome with the set of its values (literals). In fact, we also often identify such sets with the conjunctions of their elements. Sets (conjunctions) of literals corresponding to outcomes are consistent and complete.

A *conditional preference rule* (sometimes, a *preference rule* or just a *rule*) over $V$ is an expression $p : l > \neg l$, where $l$ is a literal of some atom $x \in V$ and $p$ is a propositional formula over $V$ that does not involve variable $x$.

---

[2]. In the literature one often finds the following formulation: "$Q$, even if $R$", which does not have exactly the same meaning as "$Q$, given $R$". Specifically, when saying "$Q$ is NP-complete, even if $R$", one means "$Q$ is NP-complete, *and $Q$, given $R$ is NP-complete as well*".

In the rest of the paper, we need to refer to two different languages: a conditional preference language where for every (binary) variable *x*, the conditional preference table for *x* needs to specify a preferred value of *x* for every possible assignment of its parent variables, and a more general language where the tables may be incomplete (for some values of its parents, the preferred value of *x* may not be specified) and/or locally inconsistent (for some values of its parents, the table may both contain the information that *x* is preferred and the information that ¬*x* is preferred). We call these languages respectively CP-nets and GCP-nets (for "generalized CP-nets"). Note that GCP-nets are not new, as similar structures have been discussed before (Domshlak, Rossi, Venable, & Walsh, 2003). The reason why we use this terminology ("CP-nets" and "GCP-nets") is twofold. First, even if the assumptions of completeness and local consistency for CP-nets are sometimes relaxed, most papers on CP-nets do make them. Second, we could have used "CP-nets" and "locally consistent, complete CP-nets" instead of "GCP-nets" and "CP-nets", but we felt our notation is simpler and more transparent.

**Definition 1 (Generalized CP-net)** *A generalized CP-net C (for short, a GCP-net) over V is a set of conditional preference rules. For $x \in V$ we define $p_C^+(x)$ and $p_C^-(x)$, usually written just: $p^+(x)$ and $p^-(x)$, as follows: $p_C^+(x)$ is equal to the disjunction of all p such that there exists a rule $p : x > \neg x$ in C; $p_C^-(x)$ is the disjunction of all p such that there exists a rule $p : \neg x > x$ in C. We define the associated directed graph $G_C$ (the* dependency *graph) over V to consist of all pairs $(y, x)$ of variables such that y appears in either $p^+(x)$ or $p^-(x)$.*

In our complexity results we will also need the following representation of GCP-nets: a GCP-net *C* is said to be in *conjunctive form* if *C* only contains rules $p : l > \neg l$ such that *p* is a (possibly empty) conjunction of literals. In this case all formulas $p^-(x)$, $p^+(x)$ are in disjunctive normal form, that is, a disjunction of conjunctions of literals (including ⊤ – the empty conjunction of literals).

GCP-nets determine a transitive relation on outcomes, interpreted in terms of preference. A preference rule $p : l > \neg l$ represents the statement "given that *p* holds, *l* is preferred to ¬*l* ceteris paribus". Its intended meaning is as follows. If outcome β satisfies *p* and *l*, then β is preferred to the outcome α which differs from β only in that it assigns ¬*l* to variable *x*. In this situation we say that there is *an improving flip* from α to β *sanctioned* by the rule $p : l > \neg l$.

**Definition 2** *If $\alpha_0, \ldots, \alpha_m$ is a sequence of outcomes with $m \geq 1$ and each next outcome in the sequence is obtained from the previous one by an improving flip, then we say that $\alpha_0, \ldots, \alpha_m$ is an* improving *sequence from $\alpha_0$ to $\alpha_m$ for the GCP-net, and that $\alpha_m$ dominates $\alpha_0$, written $\alpha_0 \prec \alpha_m$.*

Finally, a GCP-net is *consistent* if there is no outcome α which is strictly preferred to itself, that is, such that $\alpha \prec \alpha$.

The main objective of the paper is to establish the complexity of the following two problems concerning the notion of dominance associated with GCP-nets (sometimes under restrictions on the class of input GCP-nets).

**Definition 3**
GCP-DOMINANCE: *given a GCP-net C and two outcomes α and β, decide whether $\alpha \prec \beta$ in C, that is, whether β dominates α in C.*
GCP-CONSISTENCY: *given a GCP-net C, decide whether C is consistent.*

GCP-nets extend the notion of CP-nets (Boutilier et al., 1999). There are two properties of GCP-nets that are essential in linking the two notions.

**Definition 4**
*A GCP-net C over V is* locally consistent *if for every $x \in V$, the formula $p_C^-(x) \wedge p_C^+(x)$ is unsatisfiable. It is* locally complete *if for every $x \in V$, the formula $p_C^-(x) \vee p_C^+(x)$ is a tautology.*

Informally, local consistency means that there is no outcome in which both $x$ is preferred over $\neg x$ and $\neg x$ is preferred over $x$. Local completeness means that, for every variable $x$, in every outcome either $x$ is preferred over $\neg x$ or $\neg x$ is preferred over $x$.

**Definition 5 (Propositional CP-net)** *A* CP-net *over the set V of (propositional) variables is a locally consistent and locally complete GCP-net over V.*

It is not easy to decide whether a GCP-net is actually a CP-net. In fact, the task is coNP-complete.

**Proposition 1** *The problem of deciding, given a GCP-net C, whether C is a CP-net is* coNP-*complete.*

*Proof:* Deciding whether a GCP-net $C$ is a CP-net consists of checking local consistency and local completeness. Each of these tasks amounts to $n$ validity tests (one for each variable). It follows that deciding whether a GCP-net is a CP-net is the intersection of $2n$ problems from coNP. Hence, it is in coNP, itself. Hardness comes from the following reduction from UNSAT. To any propositional formula $\varphi$ we assign the CP-net $C(\varphi)$, defined by its set of variables $Var(\varphi) \cup \{z\}$, where $z \notin Var(\varphi)$, and the following tables:

- for any variable $x \neq z$: $p_{C(\varphi)}^+(x) = \top$; $p_{C(\varphi)}^-(x) = \bot$;

- $p_{C(\varphi)}^+(z) = \neg \varphi$; $p_{C(\varphi)}^-(z) = \bot$.

For any variable $x \neq z$, we have $p_{C(\varphi)}^+(x) \wedge p_{C(\varphi)}^-(x) = \bot$; moreover, $p_{C(\varphi)}^+(z) \wedge p_{C(\varphi)}^-(z) = \bot$. Therefore, $C(\varphi)$ is locally consistent. Now, for any variable $x \neq z$, we have $p_{C(\varphi)}^+(x) \vee p_{C(\varphi)}^-(x) = \top$. Moreover, $p_{C(\varphi)}^+(z) \vee p_{C(\varphi)}^-(z) = \neg \varphi$. Thus, $C(\varphi)$ is locally complete if and only if $\varphi$ is unsatisfiable. It follows that $C(\varphi)$ is a CP-net if and only if $\varphi$ is unsatisfiable. □

Many works on CP-nets make use of explicit conditional preference tables that list *every* combination of values of parent variables (variables on which $x$ depends) *exactly* once, each such combination designating either $x$ or $\neg x$ as preferred.[3] Clearly, CP-nets in this restricted sense can be regarded as CP-nets in our sense that, for every variable $x$, satisfy the following condition:

> if $y_1, \ldots, y_k$ are all the atoms appearing in $p^+(x)$ and $p^-(x)$ then *every* complete and consistent conjunction of literals over $\{y_1, \ldots, y_n\}$ appears as a disjunct in exactly one of $p^+(x)$ and $p^-(x)$.

---
3. There are exceptions. Some are discussed for instance by Boutilier et al. (2004a) in Section 6 of their paper.

Under this embedding, the concepts of dominance and consistency we introduced here for GCP-nets generalize the ones considered for CP-nets as defined by Boutilier et al. (2004a).

Problems CP-DOMINANCE and CP-CONSISTENCY are defined analogously to Definition 3. In the paper we are interested in the complexity of dominance and consistency problems for both GCP-nets and CP-nets. Therefore, the matter of the way in which these nets (especially CP-nets, as for GCP-nets there are no alternative proposals) are represented is important. Our representation of CP-nets is often more compact than the one proposed by Boutilier et al. (2004a), as the formulas $p^+(x)$ and $p^-(x)$ implied by the conditional preference tables can often be given equivalent, but exponentially smaller, disjunctive normal form representations. Thus, when defining a decision problem, it is critical to specify the way to represent its input instances, as the representation may affect the complexity of the problem. Unless stated otherwise, we assume that GCP-nets (and thus, CP-nets) are represented as a set of preference rules, as described in Definition 1. Therefore, the size of a GCP-net is given by the total size of the formulas $p^-(x)$, $p^+(x)$, $x \in V$.

We now note a key property of consistent GCP-nets, which we will use several times later in the paper.

**Proposition 2** *If a GCP-net C is consistent then it is locally consistent.*

*Proof:* If $C$ is not locally consistent then there exists a variable $x$ and an outcome $\alpha$ satisfying $p_C^-(x) \wedge p_C^+(x)$. Then $\alpha \prec \alpha$ can be shown by flipping $x$ from its current value in $\alpha$ to the dual value and then flipping it back: since $\alpha$ satisfies $p_C^-(x) \wedge p_C^+(x)$, and since $p_C^-(x) \wedge p_C^+(x)$ does not involve any occurrences of $x$, both flips are allowed. □

Finally, we conclude this section with an example illustrating the notions discussed above.

**Example 1** *Consider a GCP-net C on variables $V = \{x, y\}$ with four rules, defined as follows: $x : y > \neg y$; $\neg x : \neg y > y$; $y : \neg x > x$; $\neg y : x > \neg x$. We have $p^+(y) = x$, $p^-(y) = \neg x$, $p^+(x) = \neg y$ and $p^-(x) = y$. Therefore C is locally consistent and locally complete, and so is a CP-net.*

*There is a cycle of dominance between outcomes: $x \wedge y \prec \neg x \wedge y \prec \neg x \wedge \neg y \prec x \wedge \neg y \prec x \wedge y$, and so C is inconsistent. This shows that consistency is a strictly stronger property than local consistency.*

## 3. Propositional STRIPS Planning

In this section we derive some technical results on propositional STRIPS planning which form the basis of our complexity results in Sections 4 and 5. We establish the complexity of plan existence problems for propositional STRIPS planning under restrictions on input instances that make the problem of use in the studies of dominance and consistency in GCP-nets.

Let $V$ be a finite set of variables. A *state* over $V$ is a complete and consistent set of literals over $V$, which we often view as the conjunction of its members. A state is therefore equivalent to an *outcome*, defined in a CP-nets context.

**Definition 6 (Propositional STRIPS planning)** *By a* propositional STRIPS instance *we mean a tuple $\langle V, \alpha_0, \gamma, ACT \rangle$, where*

1. *V is a finite set of propositional variables;*

2. $\alpha_0$ *is a state over V, called the* initial state*;*

3. $\gamma$ *is a state called the* goal*;*[4]

4. *ACT is a finite set of* actions*, where each action $a \in ACT$ is described by a consistent conjunction of literals pre(a) (a* precondition*) and a consistent conjunction of literals post(a) (a* postcondition, or effect*).*[5]

*An action a is* executable *in a state $\alpha$ if $\alpha \models pre(a)$. The* effect *of a in state $\alpha$, denoted by $eff(a,\alpha)$, is the state $\alpha'$ containing the same literals as $\alpha$ for all variables not mentioned in post(a), and the literals of post(a). We assume that an action can be* applied *to any state, but that it does not change the state if its preconditions do not hold: if $\alpha \not\models pre(a)$ (given that states are complete, this is equivalent to $\alpha \models \neg pre(a)$) then $eff(a,\alpha) = \alpha$. This assumption has no influence as far as complexity results are concerned.*

*The* PROPOSITIONAL STRIPS PLAN EXISTENCE *problem, or* STRIPS PLAN *for short, is to decide whether for a given propositional STRIPS instance $\langle V, \alpha_0, \gamma, ACT \rangle$ there is a finite sequence of actions leading from the initial state $\alpha_0$ to the final state $\gamma$. Each such sequence is a* plan *for $\langle V, \alpha_0, \gamma, ACT \rangle$. A plan is* irreducible *if every one of its actions changes the state.*

We assume, without loss of generality, that for any action *a*, no literal in *post(a)* appears also in *pre(a)*; otherwise we can omit the literal from *post(a)* without changing the effect of the action; if *post(a)* then becomes an empty conjunction, the action *a* can be omitted from *ACT* as it has no effect.

We have the following result due to Bylander (1994).

**Proposition 3 (Bylander, 1994)** STRIPS PLAN *is* PSPACE-*complete.*

Typically, propositional STRIPS instances do not require that goals be states. Instead, goals are defined as consistent conjunctions of literals that do not need to be complete. In such a setting, a plan is a sequence of actions that leads from the start state to a state in which the goal holds. We restrict consideration to *complete* goals. This restriction has no effect on the complexity of the plan existence problem: it remains PSPACE-complete under the goal-completeness restriction (Lang, 2004).

### 3.1 Acyclic STRIPS

**Definition 7 (Acyclic sets of actions)** *A set of actions ACT (we use the same notation as in Definition 6) is* acyclic *if there is no state $\alpha$ such that $\langle V, \alpha, \alpha, ACT \rangle$ has a non-empty irreducible plan, that is to say, if there are no non-trivial directed cycles in the graph on states induced by ACT.*

We will now establish the complexity of the following problem:

ACTION-SET ACYCLICITY: given a set *ACT* of actions, decide whether *ACT* is acyclic.

**Proposition 4**
ACTION-SET ACYCLICITY *is* PSPACE-*complete.*

---

4. Note that in standard STRIPS the goal can be a partial state. This point is discussed just after Proposition 3.
5. We emphasize that we allow negative literals in preconditions and goals. Some definitions of STRIPS do not allow this. This particular variant of STRIPS is sometimes called PSN (propositional STRIPS with negation) in the literature.

*Proof:* The argument for the membership in PSPACE is standard; we nevertheless give some details. We will omit such details for further proofs of membership in PSPACE. The following nondeterministic algorithm decides that *ACT* has a cycle:

guess $\alpha_0$;
$\alpha := \alpha_0$;
**repeat**
    guess an action $a \in ACT$;
    $\alpha' := \textit{eff}(a, \alpha)$;
    $\alpha := \alpha'$
**until** $\alpha = \alpha_0$.

This algorithm works in nondeterministic polynomial space (because we only need to store $\alpha_0$, $\alpha$ and $\alpha'$), which shows that $\overline{\text{ACTION-SET ACYCLICITY}}$ is in NPSPACE, and therefore in PSPACE, since NPSPACE = PSPACE. Thus, ACTION-SET ACYCLICITY is in coPSPACE, hence in PSPACE, since coPSPACE = PSPACE.

We will now show that the *complement* of the ACTION-SET ACYCLICITY problem is PSPACE-hard by reducing the ACYCLIC STRIPS PLAN problem to it.

Let $PE = \langle V, \alpha_0, \gamma, ACT \rangle$ be an instance of the ACYCLIC STRIPS PLAN problem. In particular, we have that *ACT* is acyclic. Let *a* be a new action defined by $pre(a) = \gamma$ and $post(a) = \alpha_0$. It is easy to see that $ACT \cup \{a\}$ is *not* acyclic if and only if there exists a plan for *PE*. Thus, the PSPACE-hardness of the *complement* of the ACTION-SET ACYCLICITY problem follows from Proposition 5. Consequently, the ACTION-SET ACYCLICITY problem is coPSPACE-hard. Since PSPACE = coPSPACE, the ACTION-SET ACYCLICITY problem is PSPACE-hard, as well. □

Next, we consider the STRIPS planning problem restricted to instances that have acyclic sets of actions. Formally, we consider the following problem:

> ACYCLIC STRIPS PLAN: Given a propositional STRIPS instance $\langle V, \alpha_0, \gamma, ACT \rangle$ such that *ACT* is acyclic and $\alpha_0 \neq \gamma$, decide whether there is a plan for $\langle V, \alpha_0, \gamma, ACT \rangle$

This is the first of our problems of the form "*Q*, given *R*" that we encounter and it illustrates well the concerns we discussed at the end of the introduction. Here, *R* is the set of all propositional STRIPS instances $\langle V, \alpha_0, \gamma, ACT \rangle$ such that *ACT* is acyclic, and *Q* is the set of all such instances for which there is a plan for $\langle V, \alpha_0, \gamma, ACT \rangle$. Checking whether a given propositional STRIPS instance is actually acyclic is itself PSPACE-complete (this is what Proposition 4 states), but this does not matter when it comes to solving ACYCLIC STRIPS PLAN: when considering an instance of ACYCLIC STRIPS PLAN, we already know that it is acyclic (and this is reflected in the reduction below).

**Proposition 5**
ACYCLIC STRIPS PLAN *is* PSPACE-*complete.*

*Proof:* The argument for the membership in PSPACE is standard (cf. the proof of Proposition 4). To prove PSPACE-hardness, we first exhibit a polynomial-time reduction *F* from STRIPS PLAN. Let $PE = \langle V, \alpha_0, \gamma, ACT \rangle$ be an instance of STRIPS PLAN. The idea behind the reduction is to introduce a *counter*, so that each time an action is executed, the counter is incremented. The counter may count up to $2^n$, where $n = |V|$, making use of *n* additional variables. The counter is initialized to

0. Once it reaches $2^n - 1$ it can no longer be incremented and no action can be executed. Hence, the set of actions in the resulting instance of STRIPS PLAN is acyclic: we are guaranteed to produce an instance of $R$. To describe the reduction, we write $V$ as $\{x_1, \ldots, x_n\}$. We define $F(PE) = PE' = \langle V', \alpha'_0, \gamma', ACT' \rangle$ as follows:

- $V' = \{x_1, \ldots, x_n, z_1, \ldots, z_n\}$, where $z_i$ are new variables we will use to implement the counter;

- $\alpha'_0 = \alpha_0 \wedge \neg z_1 \wedge \cdots \wedge \neg z_n$;

- $\gamma' = \gamma \wedge z_1 \wedge \cdots \wedge z_n$;

- for each action $a \in ACT$, we include in $ACT'$ $n$ actions $a^i$, $1 \leq i \leq n$, such that:

  - for $i \leq n-1$: $\begin{cases} pre(a^i) = pre(a) \wedge \neg z_i \wedge z_{i+1} \wedge \cdots \wedge z_n \\ post(a^i) = post(a) \wedge z_i \wedge \neg z_{i+1} \wedge \cdots \wedge \neg z_n, \text{ and} \end{cases}$

  - for $i = n$: $\begin{cases} pre(a^n) = pre(a) \wedge \neg z_n \\ post(a^n) = post(a) \wedge z_n. \end{cases}$

- Furthermore, we include in $ACT'$ $n$ actions $b_i$, $1 \leq i \leq n$, such that:

  - for $i \leq n-1$: $\begin{cases} pre(b^i) = \neg z_i \wedge z_{i+1} \wedge \cdots \wedge z_n \\ post(b^i) = z_i \wedge \neg z_{i+1} \wedge \cdots \wedge \neg z_n, \text{ and} \end{cases}$

  - for $i = n$: $\begin{cases} pre(b^n) = \neg z_n \\ post(b^n) = z_n. \end{cases}$

We will denote states over $V'$ by pairs $(\alpha, k)$, where $\alpha$ is a state over $V$ and $k$ is an integer, $0 \leq k \leq 2^n - 1$. We view $k$ as a compact representation of a state over variables $z_1, \ldots, z_n$: assuming that the binary representation of $k$ is $d_1 \ldots d_n$ (with $d_n$ being the least significant digit), $k$ represents the state which contains $z_i$ if $d_i = 1$ and $\neg z_i$, otherwise. For instance, let $V = \{x_1, x_2, x_3\}$. Then we have $V' = \{x_1, x_2, x_3, z_1, z_2, z_3\}$, and the state $\neg x_1 \wedge x_2 \wedge x_3 \wedge z_1 \wedge \neg z_2 \wedge z_3$ is denoted by $(\neg x_1 \wedge x_2 \wedge x_3, 5)$.

We note that the effect of $a^i$ or $b^i$ on state $(\alpha, k)$ is either void, or increments the counter:

$$eff(a^i, (\alpha, k)) = \begin{cases} (eff(a, \alpha), k+1) & \text{if } a^i \text{ is executable in } (\alpha, k) \\ (\alpha, k) & \text{otherwise} \end{cases}$$

$$eff(b^i, (\alpha, k)) = \begin{cases} (\alpha, k+1) & \text{if } b^i \text{ is executable in } (\alpha, k) \\ (\alpha, k) & \text{otherwise} \end{cases}$$

Next, we remark that at most one $a^i$ and at most one $b^i$ are executable in a given state $(\alpha, k)$. More precisely,

- if $k < 2^n - 1$, then exactly one $b^i$ is executable in $(\alpha, k)$; denote by $i(k)$ the index such that $b^{i(k)}$ is executable in $(\alpha, k)$ (this index depends only on $k$). We also have that $a^{i(k)}$ is executable in $(\alpha, k)$, provided that $a$ is executable in $\alpha$.

- if $k = 2^n - 1$, then no $a^i$ and no $b^i$ is executable in $(\alpha, k)$.

Now we show that $PE'$ is acyclic. Assume $\pi$ is an irreducible plan for $\langle V', \alpha', \alpha', ACT'\rangle$. Let $\alpha' = (\alpha, k)$. If $k < 2^n - 1$, then $\pi$ is empty, since any action in $ACT'$ in any state either is non-executable or increments the counter, and an irreducible plan contains only actions whose effect is non-void. If $k = 2^n - 1$, then no action of $ACT'$ is executable in $\alpha'$ and again $\pi$ is empty. Thus, there exists no non-empty irreducible plan for $\langle V', \alpha', \alpha', ACT'\rangle$, and this holds for all $\alpha'$. Therefore $PE'$ is acyclic.

We now claim that there is a plan for $PE$ if and only if there is a plan for $PE'$. First, assume that there is a plan in $PE$. Let $\pi$ be a shortest plan in $PE$ and let $m$ be its length (the number of actions used). We have $m \leq 2^n - 1$, since no state along $\pi$ repeats (otherwise, shorter plans than $\pi$ for $PE$ would exist). Let $\alpha_0, \alpha_1, \ldots, \alpha_m = \gamma$ be the sequence of states obtained by executing $\pi$. Let $a$ be the action used in the transition from $\alpha_k$ to $\alpha_{k+1}$. Since $k < 2^n - 1$ (because $m \leq 2^n - 1$ and $k \leq m - 1$), there is exactly one $i$, $1 \leq i \leq n$, such that the action $a^i$ applies at the state $(\alpha, k)$ over $V'$. Replacing $a$ with $a^i$ in $\pi$ yields a plan that when started at $(\alpha_0, 0)$ leads to $(\alpha_m, m) = (\gamma, m)$. Appending that plan with appropriate actions $b_i$ to increment the counter to $2^n - 1$ yields a plan for $PE'$. Conversely, if $\tau$ is a plan for $PE'$, the plan obtained from $\tau$ by removing all actions of the form $b^j$ and replacing each action $a^i$ with $a$ is a plan for $PE$, since $a^i$ has the same effect on $V$ as $a$ does. Thus, the claim follows. □

We emphasize that this reduction $F$ from STRIPS PLAN to ACYCLIC STRIPS PLAN (or, equivalently, to STRIPS PLAN given ACTION-SET ACYCLICITY) works because it satisfies the following two conditions:

1. for every instance $PE$ of STRIPS PLAN, $F(PE)$ is an instance of ACYCLIC STRIPS PLAN (this holds because for every $PE$, $F(PE)$ is acyclic);

2. for every $PE$ of STRIPS PLAN, $F(PE)$ is a positive instance of ACYCLIC STRIPS PLAN if and only if $PE$ is a positive instance of STRIPS PLAN.

### 3.2 Mapping STRIPS Plans to Single-Effect STRIPS Plans

Versions of the STRIPS PLAN and ACYCLIC STRIPS PLAN problems that are important for us allow only actions with exactly one literal in their postconditions in their input propositional STRIPS instances. We call such actions *single-effect actions*.[6] We refer to the restricted problems as SE STRIPS PLAN and ACYCLIC SE STRIPS PLAN, respectively.

To prove PSPACE-hardness of both problems, we describe a mapping from STRIPS instances to single-effect STRIPS instances.[7]

Consider an instance $PE = \langle V, \alpha_0, \gamma, ACT\rangle$ of the STRIPS PLAN problem, where $ACT$ is not necessarily acyclic. For each action $a \in ACT$ we introduce a *new* variable $x_a$, whose intuitive meaning is that action $a$ is currently being executed.

We set $X = \bigwedge_{a \in ACT} \neg x_a$. That is, $X$ is the conjunction of negative literals of all the additional variables. In addition, for each $a \in ACT$ we set $X_a = x_a \wedge \bigwedge_{b \in ACT - \{a\}} \neg x_b$. We now define an instance $PE' = \langle V', \alpha'_0, \gamma', S(ACT)\rangle$ of the SE STRIPS PLAN problem as follows:

---

6. Such actions are also called "unary" actions in the planning literature. We stick to the terminology "single-effect" although it is less commonly used, simply because it is more explicit.
7. PSPACE-completeness of propositional STRIPS planning with single-effect actions was proved already by Bylander (1994). However, to deal with acyclicity we need to give a different reduction than the one used in that paper.

- Set of variables: $V' = V \cup \{x_a : a \in ACT\}$;
- initial state: $\alpha'_0 = \alpha_0 \wedge X$;
- goal state: $\gamma' = \gamma \wedge X$;
- set of actions: $S(ACT) = \{a^i : a \in ACT, i = 1, \ldots, 2|post(a)| + 1\}$.
  Let $a$ be an action in $ACT$ such that $post(a) = l_1 \wedge \cdots \wedge l_q$, where $l_1, \ldots, l_q$ are literals.
    - For $i = 1, \ldots, q$, we define an action $a^i$ by setting:
    
    $$pre(a^i) = pre(a) \wedge X \wedge \neg l_i; \ post(a^i) = x_a.$$
    
    The role of $a^i$ is to enforce that $X_a$ holds after $a^i$ is successfully applied, and in this way to enable "starting the execution of $a$", provided that no action is currently being executed, that the $i$th effect of $a$ is not already true, and that the precondition of $a$ is true.
    - For $i = q+1, \ldots, 2q$, we define action $a^i$ by setting:
    
    $$pre(a^i) = X_a; \ post(a^i) = l_i.$$
    
    The role of $a^i$ is to make the $i$th effect of $a$ true.
    - Finally, we define $a^{2q+1}$ by setting:
    
    $$pre(a^{2q+1}) = X_a \wedge l_1 \wedge \cdots \wedge l_q; \ post(a^{2q+1}) = \neg x_a.$$
    
    Thus, $a^{2q+1}$ is designed so that $X$ holds after $a^{2q+1}$ is successfully applied; that is, $a^{2q+1}$ "closes" the execution of $a$, thus allowing for the next action to be executed.

Let $\pi$ be a sequence of actions in $ACT$. We define $S(\pi)$ to be the sequence of actions in $S(ACT)$ obtained by replacing each action $a$ in $\pi$ by $a^1, \ldots, a^{2q+1}$, where $q = |post(a)|$. Now consider a sequence $\tau$ of actions from $S(ACT)$. Remove from $\tau$ every action $a^i$ such that $i \neq 2|post(a)| + 1$, and replace actions of the form $a^{2|post(a)|+1}$ by $a$. We denote the resulting sequence of actions from $ACT$ by $S'(\tau)$. We note that $S'(S(\pi)) = \pi$. The following properties then hold.

**Lemma 1** *With the above definitions,*

(i) *if $\pi$ is a plan for PE then $S(\pi)$ is a plan for PE';*

(ii) *if $\tau$ is an irreducible plan for PE' then $S'(\tau)$ is an irreducible plan for PE;*

(iii) *ACT is acyclic if and only if $S(ACT)$ is acyclic.*

*Proof:* (i) Let $a \in ACT$ be an action, let $\alpha$ be a state and let $\beta$ be the state obtained from $\alpha$ by applying $a$. Let $\theta$ be the $V'$-state obtained by applying the sequence of actions $\langle a^1, \ldots, a^{2q+1} \rangle$ (where $q = |post(a)|$) to the state $\alpha \wedge X$ of $PE'$. We will show that $\theta = \beta \wedge X$.

We note that if for each $i = 1, \ldots, q$, state $\alpha \wedge X$ does not satisfy $pre(a^i)$ then the sequence of actions $\langle a^1, \ldots, a^{2q+1} \rangle$ has no effect, so the state is still $\alpha \wedge X$. For this to happen, either $\alpha$ doesn't satisfy $pre(a)$, or all of $l_1, \ldots, l_q$ already hold in $\alpha$ so $post(a)$ holds in $\alpha$. In either case, $\alpha = \beta$, and so $\theta = \beta \wedge X$.

Suppose now that for some $i \in \{1, \ldots, q\}$, $\alpha$ does satisfy $pre(a^i)$. Then the first such action $a^i$ causes $x_a$ and hence $X_a$ to hold. After applying actions $a^{q+1}, \ldots, a^{2q}$, $l_1 \wedge \cdots \wedge l_q$ holds, and so $post(a)$ holds. After applying $a^{2q+1}$ both $post(a)$ and $X$ hold. No other variable in $V$ has changed, so $\theta = \beta \wedge X$, as required.

Applying this result iteratively implies that if $\pi$ is a plan for $PE$ then $S(\pi)$ is a plan for $PE'$.

(ii) Let $\tau$ be an irreducible plan for $PE'$, so that every action in $\tau$ changes the state, which implies that every action in $\tau$ is performed in a state where its precondition is true. We will show that $S'(\tau)$ is a plan for $PE$. We will assume that $\tau \neq \emptyset$. When $\tau = \emptyset$, $S'(\tau) = \emptyset$, too, and the assertion follows.

Write the first action in $\tau$ as $a^j$, where $a \in ACT$, and let $\tau'$ be the maximal initial subsequence of $\tau$ consisting of all actions of the form $a^i$. We must have $j \leq |post(a)|$, since $X$ holds in $\alpha_0'$ (by our assumption above, action $a^j$ does apply) and $X$ is inconsistent with the precondition of $a^i$ for each $i > |post(a)|$. Also, $pre(a^j)$ and $\neg l_j$ hold in $\alpha_0'$ and so, in $\alpha_0$ as well. Thus, $\alpha_0$ satisfies $pre(a)$, and applying $a$ changes the state, since $\neg l_j$ holds in $\alpha_0$ and $post(a) \models l_j$. Let us denote by $\beta$ the state resulting from applying $a$ to $\alpha_0$. As we noted, $\beta \neq \alpha_0$,

Let $\beta'$ be the state resulting after applying $\tau'$ to $\alpha_0'$. If $\beta'$ is the goal state $\gamma'$ then $X$ holds in $\beta'$. If $\beta'$ is not the goal state then $\tau \neq \tau'$. Let $b^i$ be the action in $\tau$ directly following the last action in $\tau'$. By the definition of $\tau'$, $a \neq b$. After applying $a^j$, $X_a$ holds, so in $\beta'$ either $X_a$ holds or $X$ holds. Thus, $X_b$ does not hold, as $a \neq b$. Since $b^i$ changes the state, $i$ must be in $\{1, \ldots, |post(b)|\}$, so $X$ holds in $\beta'$ in this case, too.

Hence the last action in $\tau'$ is $a^{2q+1}$, where $q = |post(a)|$. Since the only variables in $V$ which can be affected by actions $a^i$ are those that appear in the literals in $post(a)$ and since the action $a^{2q+1}$ can be executed (otherwise it would not belong to $\tau$), it follows that $\beta' = \beta \wedge X$.

Applying this reasoning repeatedly, we show that applying $S'(\tau)$ to $\alpha_0$ yields $\gamma$, and that each action in $S'(\tau)$ changes the state, so $S'(\tau)$ is an irreducible plan for $PE$, which is non-empty if and only if $\tau$ is non-empty.

(iii) Suppose $ACT$ is not acyclic, so that there exists state $\alpha$ and a non-empty irreducible plan $\pi$ for $PE_\alpha = \langle V, \alpha, \alpha, ACT \rangle$. Then, by (i), $S(\pi)$ is a plan for $PE_\alpha' = \langle V', \alpha \wedge X, \alpha \wedge X, S(ACT) \rangle$. Because $\pi$ is non-empty and irreducible, it changes some state, so $S(\pi)$ also changes some state, and hence can be reduced to a non-empty irreducible plan for $PE_\alpha'$. Therefore $S(ACT)$ is not acyclic.

Conversely, suppose that $S(ACT)$ is not acyclic. Then there exists a state $\alpha'$ and a non-empty irreducible plan $\tau$ for $\langle V', \alpha', \alpha', S(ACT) \rangle$. We will first prove that $X$ holds at some state obtained during the execution of this plan.

Suppose that $X$ holds at no such state, and let $a^j$ be the first action in $\tau$. We note that $\tau \neq \emptyset$. By our assumption, $X$ does not hold either before or after applying $a_j$. Therefore $q + 1 \leq j \leq 2q$, where $q = |post(a)|$. Since $\tau$ is irreducible, $a^j$ changes the state. Thus, $\neg l_j$ holds in $\alpha'$ and $l_j$ holds in the state resulting from $\alpha'$ after applying $a^j$.

By our assumption, $X_a$ holds before and after applying $a_j$. Thus, the next action, if there is one, must also be of the form $a^i$ for $q + 1 \leq i \leq 2q$. Repeating this argument implies that all actions in $\tau$ are of the form $a^i$ where $q + 1 \leq i \leq 2q$. Since the set of literals in $post(a)$ is consistent, $l_j$ is never reset back to $\neg l_j$. Thus, the state resulting from $\alpha'$ after applying $\tau$ is different from $\alpha'$, a contradiction.

Thus, $X$ holds at some state reached during the execution of $\tau$. Let us consider one such state. It can be written as $\beta \wedge X$, for some state $\beta$ over $V$. We can cyclically permute $\tau$ to generate a non-empty irreducible plan $\tau'$ for $\langle V', \beta \wedge X, \beta \wedge X, S(ACT) \rangle$. By part (ii), $S'(\tau')$ is a non-empty irreducible plan for $\langle V, \beta, \beta, ACT \rangle$. Therefore $ACT$ is not acyclic. □

**Proposition 6**
SE STRIPS PLAN *and* ACYCLIC SE STRIPS PLAN *are* PSPACE-*complete.*

*Proof:* Again, the argument for the membership in PSPACE is standard. PSPACE-hardness of ACYCLIC SE STRIPS PLAN is shown by reduction from ACYCLIC STRIPS PLAN. The same construction shows that STRIPS PLAN is reducible to SE STRIPS PLAN, and thus SE STRIPS PLAN is PSPACE-complete.

Let us consider an instance $PE = \langle V, \alpha_0, \gamma, ACT \rangle$ of ACYCLIC STRIPS PLAN. We define $PE' = \langle V', \alpha'_0, \gamma', S(ACT) \rangle$, which by Lemma 1(iii) is an instance of the ACYCLIC SE STRIPS PLAN problem. By Lemma 1(i) and (ii) there exists a plan for $PE$ if and only if there exists a plan for $PE'$. This implies that ACYCLIC SE STRIPS PLAN is PSPACE-hard. □

## 4. Dominance

The goal of this section is to prove that the GCP-DOMINANCE problem is PSPACE-complete, and that the complexity does not go down even when we restrict the class of inputs to CP-nets. We use the results on propositional STRIPS planning from Section 3 to prove that the general GCP-DOMINANCE problem is PSPACE-complete. We then show that the complexity does not change if we require the input GCP-net to be locally consistent and locally complete.

The similarities between dominance testing in CP-nets and propositional STRIPS planning were first noted by Boutilier et al. (1999). They presented a reduction, discussed later in more detail by Boutilier et al. (2004a), from the dominance problem to the plan existence problem for a class of propositional STRIPS planning specifications consisting of *unary* actions (actions with single effects). We prove our results for the GCP-DOMINANCE and GCP-CONSISTENCY problems by constructing a reduction in the other direction.

This reduction is much more complex than the one used by Boutilier et al. (1999), due to the fact that CP-nets impose more restrictions than STRIPS planning. Firstly, STRIPS planning allows multiple effects, but GCP-nets only allow flips $x > \neg x$ or $\neg x > x$ that change the value of one variable; this is why we constructed the reduction from STRIPS planning to single-effect STRIPS planning in the last section. Secondly, CP-nets impose two more restrictions, local consistency and local completeness, which do not have natural counterparts in the context of STRIPS planning.

For all dominance and consistency problems we consider, the membership in PSPACE can be demonstrated similarly to the membership proof of Proposition 4, namely by considering nondeterministic polynomial space algorithms consisting of repeatedly guessing appropriate improving flips and making use of the fact that PSPACE = NPSPACE = coPSPACE. Therefore, from now on we only provide arguments for the PSPACE-hardness of problems we consider.

### 4.1 Dominance for Generalized CP-Nets

We will prove that the GCP-DOMINANCE problem is PSPACE-complete by a reduction from the problem SE STRIPS PLAN, which we now know to be PSPACE-complete.

### 4.1.1 MAPPING SINGLE-EFFECT STRIPS PROBLEMS TO GCP-NETS DOMINANCE PROBLEMS

Let $\langle V, \alpha_0, \gamma, ACT \rangle$ be an instance of the SE STRIPS PLAN problem. For every action $a \in ACT$ we denote by $l_a$ the unique literal in the postcondition of $a$, that is, $post(a) = l_a$. We denote by $pre'(a)$ the conjunction of all literals in $pre(a)$ different from $\neg l_a$ (we recall that by a convention we adopted earlier, $pre'(a)$ does not contain $l_a$). We then define $c_a$ to be the conditional preference rule $pre'(a) : l_a > \neg l_a$ and define $M(ACT)$ to be the GCP-net $C = \{c_a : a \in ACT\}$, which is in conjunctive form.

A sequence of states in a plan corresponds to an improving sequence from $\alpha_0$ to $\gamma$, which leads to the following result.

**Lemma 2** *With the above notation,*

(i) *there is a non-empty irreducible plan for $\langle V, \alpha_0, \gamma, ACT \rangle$ if and only if $\gamma$ dominates $\alpha_0$ in $M(ACT)$;*

(ii) *ACT is acyclic if and only if $M(ACT)$ is consistent.*

*Proof:* We first note the following equivalence. Let $a$ be an action in $ACT$, and let $\alpha$ and $\beta$ be different outcomes (or, in the STRIPS setting, states). The action $a$ applied to $\alpha$ yields $\beta$ if and only if the rule $c_a$ sanctions an improving flip from $\alpha$ to $\beta$. This is because $a$ applied to $\alpha$ yields $\beta$ if and only if $\alpha$ satisfies $pre(a)$ and $\alpha$ and $\beta$ differ only on literal $l_a$, with $\beta$ satisfying $l_a$ and $\alpha$ satisfying $\neg l_a$. This is if and only if $\alpha$ satisfies $pre'(a)$ and $\alpha$ and $\beta$ differ only on literal $l_a$, with $\beta$ satisfying $l_a$, and $\alpha$ satisfying $\neg l_a$. This, in turn, is equivalent to say that rule $c_a$ sanctions an improving flip from $\alpha$ to $\beta$.

Proof of (i): Suppose first that there exists a non-empty irreducible plan $a_1, \ldots, a_m$ for $\langle V, \alpha_0, \gamma, ACT \rangle$. Let $\alpha_0, \alpha_1, \ldots, \alpha_m = \gamma$ be the corresponding sequence of outcomes, and, for each $i = 1, \ldots, m$, action $a_i$, when applied in state $\alpha_{i-1}$, yields different state $\alpha_i$. By the above equivalence, for each $i = 1, \ldots, m$, $c_{a_i}$ sanctions an improving flip from $\alpha_{i-1}$ to $\alpha_i$, which implies that $\alpha_0, \alpha_1, \ldots, \alpha_m$ is an improving flipping sequence in $M(ACT)$, and therefore $\gamma$ dominates $\alpha_0$ in $M(ACT)$.

Conversely, suppose that $\gamma$ dominates $\alpha_0$ in $M(ACT)$, so that there exists an improving flipping sequence $\alpha_0, \alpha_1, \ldots, \alpha_m$ with $\alpha_m = \gamma$, and $m \geq 1$. For each $i = 1, \ldots, m$, let $c_{a_i}$ be an element of $M(ACT)$ which sanctions the improving flip from $\alpha_{i-1}$ to $\alpha_i$. Then, by the above equivalence, action $a_i$, when applied to state $\alpha_{i-1}$ yields $\alpha_i$ (which is different from $\alpha_{i-1}$), and so $a_1, \ldots, a_m$ is a non-empty irreducible plan for $\langle V, \alpha_0, \gamma, ACT \rangle$.

Proof of (ii): *ACT* is *not* acyclic if and only if there exists a state $\alpha$ and a non-empty irreducible plan for $\langle V, \alpha, \alpha, ACT \rangle$. By (i) this is if and only if there exists an outcome $\alpha$ which dominates itself in $M(ACT)$, which is if and only if $M(ACT)$ is not consistent. □

**Theorem 1** *The GCP-DOMINANCE problem is PSPACE-complete. Moreover, this remains so under the restrictions that the GCP-net is consistent and is in conjunctive form.*

*Proof:* PSPACE-hardness is shown by reduction from ACYCLIC SE STRIPS PLAN (Proposition 6). Let $\langle V, \alpha_0, \gamma, ACT \rangle$ be an instance of the ACYCLIC SE STRIPS PLAN problem. By Lemma 2(ii), $M(ACT)$ is a consistent GCP-net in conjunctive form. Since $\alpha_0 \neq \gamma$ (imposed in the definition of the problem ACYCLIC SE STRIPS PLAN), there is a plan for $\langle V, \alpha_0, \gamma, ACT \rangle$ if and only if there is a non-empty irreducible plan for $\langle V, \alpha_0, \gamma, ACT \rangle$, which, by Lemma 2(i), is if and only if $\gamma$ dominates $\alpha_0$ in $C$. □

Theorem 1 implies the PSPACE-completeness of dominance in the more general conditional preference language introduced by Wilson (2004b), where the conditional preference rules are written in conjunctive form.

### 4.2 Dominance in CP-Nets

In this section we show that GCP-DOMINANCE remains PSPACE-complete under the restriction to locally consistent and locally complete GCP-nets, that is, CP-nets. We refer to this restriction of GCP-DOMINANCE as CP-DOMINANCE.

Consistency of a GCP-net implies local consistency (Proposition 2). Therefore, the reduction in the proof of Theorem 1 (from ACYCLIC SE STRIPS PLAN to GCP-DOMINANCE *restricted to consistent GCP-nets*) is also a reduction to GCP-DOMINANCE restricted to locally consistent GCP-nets. PSPACE-hardness of ACYCLIC SE STRIPS PLAN (Proposition 6) then implies that GCP-DOMINANCE restricted to locally consistent GCP-nets is PSPACE-hard, and, in fact, PSPACE-complete since membership in PSPACE is easily obtained with the usual line of argumentation.

We will show PSPACE-hardness for CP-DOMINANCE by a reduction from GCP-DOMINANCE for consistent GCP-nets.

#### 4.2.1 MAPPING LOCALLY CONSISTENT GCP-NETS TO CP-NETS

Let $C$ be a locally consistent GCP-net. Let $V = \{x_1, \ldots, x_n\}$ be the set of variables of $C$. We define $V' = V \cup \{y_1, \ldots, y_n\}$, where $\{y_1, \ldots, y_n\} \cap V = \emptyset$. We define a GCP-net $C'$ over $V'$, which we will show is a CP-net. To this end, for every $z \in V'$ we will define conditional preference rules $q^+(z): z > \neg z$ and $q^-(z): \neg z > z$ to be included in $C'$ by specifying formulas $q^+(z)$ and $q^-(z)$.

First, for each variable $x_i \in V$, we set

$$q^+(x_i) = y_i \quad \text{and} \quad q^-(x_i) = \neg y_i.$$

Thus, $x_i$ depends only on $y_i$. We also note that the formulas $q^+(x_i)$ and $q^-(x_i)$ satisfy local consistency and local completeness requirements.

Next, for each variable $y_i$, $1 \leq i \leq n$, we define

$$e_i = (x_1 \leftrightarrow y_1) \wedge \cdots \wedge (x_{i-1} \leftrightarrow y_{i-1}) \wedge (x_{i+1} \leftrightarrow y_{i+1}) \wedge \cdots \wedge (x_n \leftrightarrow y_n),$$

$$f_i^+ = e_i \wedge p^+(x_i) \quad \text{and} \quad f_i^- = e_i \wedge p^-(x_i).$$

Finally, we define

$$q^+(y_i) = f_i^+ \vee (\neg f_i^- \wedge x_i)$$

and

$$q^-(y_i) = f_i^- \vee (\neg f_i^+ \wedge \neg x_i).$$

Thus, $y_i$ depends on every variable in $V'$ but itself.

We note that by the local consistency of $C$, formulas $f_i^+ \wedge f_i^-$, $1 \leq i \leq n$, are unsatisfiable. Consequently, formulas $q^+(y_i) \wedge q^-(y_i)$, $1 \leq i \leq n$, are unsatisfiable. Thus, $C'$ is locally consistent.

Finally, $q^+(y_i) \vee q^-(y_i)$ is equivalent to $f_i^+ \vee \neg x_i \vee f_i^- \vee x_i$, so is a tautology. Thus, $C'$ is locally complete and hence a CP-net over $V'$.

Let $\alpha$ and $\beta$ be outcomes over $\{x_1,\ldots,x_n\}$ and $\{y_1,\ldots,y_n\}$, respectively. By $\alpha\beta$ we denote the outcome over $V'$ obtained by concatenating $n$-tuples $\alpha$ and $\beta$. Conversely, every outcome for $C'$ can be written in this way.

Let $\alpha$ be an outcome over $V$. We define $\overline{\alpha}$ to be the outcome over $\{y_1,\ldots,y_n\}$ obtained by replacing in $\alpha$ every component of the form $x_i$ with $y_i$ and every component $\neg x_i$ with $\neg y_i$. Then for every $i$, $1 \leq i \leq n$, $\alpha\overline{\alpha} \models e_i$.

Let $s$ be a sequence $\alpha_0,\ldots,\alpha_m$ of outcomes over $V$. Define $L(s)$ to be the sequence of $V'$-outcomes: $\alpha_0\overline{\alpha_0}, \alpha_0\overline{\alpha_1}, \alpha_1\overline{\alpha_1}, \alpha_1\overline{\alpha_2},\ldots,\alpha_m\overline{\alpha_m}$. Further, let $t$ be a sequence $\varepsilon_0,\varepsilon_1,\ldots,\varepsilon_m$ of $V'$-outcomes with $\varepsilon_0 = \alpha\overline{\alpha}$ and $\varepsilon_m = \beta\overline{\beta}$. Define $L'(t)$ to be the sequence obtained from $t$ by projecting each element in $t$ to $V$ and iteratively removing elements in the sequence which are the same as their predecessor (until any two consecutive outcomes are different).

**Lemma 3** *With the above definitions,*

*(i) if $s$ is an improving sequence for $C$ from $\alpha$ to $\beta$ then $L(s)$ is an improving sequence for $C'$ from $\alpha\overline{\alpha}$ to $\beta\overline{\beta}$;*

*(ii) if $t$ is an improving sequence from $\alpha\overline{\alpha}$ to $\beta\overline{\beta}$ then $L'(t)$ is an improving sequence from $\alpha$ to $\beta$;*

*(iii) $C$ is consistent if and only if $C'$ is consistent.*

*Proof:* Let $e = \bigwedge_{i=1}^{n}(x_i \leftrightarrow y_i)$. The definitions have been arranged so that the GCP-net $C$ and the CP-net $C'$ have the following properties:
(a) If $e$ does not hold in an outcome $\gamma$ over $V'$, then every improving flip applicable to $\gamma$ changes the value of some variable $x_i$ or $y_i$ so that $x_i \leftrightarrow y_i$ holds after the flip.

Indeed, let us assume that there is an improving flip from $\gamma$ to some outcome $\gamma'$ over $V'$. If the flip concerns a variable $x_i$, then $x_i \leftrightarrow \neg y_i$ holds in $\gamma$. Consequently, $x_i \leftrightarrow y_i$ holds in $\gamma'$.

Thus, let us assume that the flip concerns a variable $y_i$. If $e_i$ holds in $\gamma$ then, since $e$ does not, $x_i \leftrightarrow \neg y_i$ holds in $\gamma$. Thus, $x_i \leftrightarrow y_i$ holds in $\gamma'$. If $e_i$ does not hold in $\gamma$ then neither $f_i^+$ nor $f_i^-$ does. Thus, if $x_i$ ($\neg x_i$, respectively) holds in $\gamma$, $y_i$ ($\neg y_i$, respectively) holds in $\gamma'$. Since the flip concerns $y_i$, it follows that $x_i \leftrightarrow y_i$ holds in $\gamma'$.
(b) No improving flip from $\alpha\overline{\alpha}$ changes any variable $x_i$.

Indeed, for any variable $x_i$, since $e$ holds in $\alpha\overline{\alpha}$, $x_i \leftrightarrow y_i$ holds in $\alpha\overline{\alpha}$, too. Thus, no improving flip changes $x_i$.
(c) There is an improving flip in $C'$ that changes variable $y_i$ in an outcome $\alpha\overline{\alpha}$ if and only if there is an improving flip for the GCP-net $C$ from outcome $\alpha$ that changes variable $x_i$. After applying the improving flip (changing variable $y_i$) to $\alpha\overline{\alpha}$, there is exactly one improving flip possible. It changes $x_i$ and results in an outcome $\beta\overline{\beta}$, where $\beta$ is the outcome over $V$ resulting from applying to $\alpha$ the improving flip changing the variable $x_i$.

To prove (c), let us first assume that $\neg y_i$ holds in $\alpha\overline{\alpha}$ and observe that in such case $\neg x_i$ holds in $\alpha\overline{\alpha}$, too. It follows that $q^+(y_i)$ holds in $\alpha\overline{\alpha}$ if and only if $p^+(x_i)$ holds in $\alpha$. Consequently, changing $y_i$ in $\alpha\overline{\alpha}$ is an improving flip in $C'$ if and only if changing $x_i$ in $\alpha$ is an improving flip in $C$. The argument in the case when $y_i$ holds in $\alpha\overline{\alpha}$ is analogous (but involves $q^-(y_i)$ and $p^-(x_i)$). Thus, the first part of (c) follows.

Let $\beta$ be the outcome obtained by applying an improving flip to $x_i$ in $\alpha$. It follows that the improving flip changing the value of $y_i$ in $\alpha\bar{\alpha}$ results in the outcome $\alpha\bar{\beta}$. In this outcome, by (a), an improving flip must concern $x_j$ or $y_j$ such that $x_j \leftrightarrow y_j$ holds after the flip. Since for every $j \neq i$, $x_j \leftrightarrow y_j$ holds in $\alpha\bar{\beta}$, the only improving flips in $\alpha\bar{\beta}$ concern either $x_i$ or $y_i$. By the local consistency of $C'$, $y_i$ cannot be flipped right back. Clearly, changing $x_i$ is an improving flip that can be applied to $\alpha\bar{\beta}$. By our discussion, it is the *only* improving flip applicable in $\alpha\bar{\beta}$ and it results in the outcome $\beta\bar{\beta}$. This proves the second part of (c).

Proof of (i): The assertion follows by iterative application of (c).

Proof of (ii): Suppose that $t$ is an improving sequence $\varepsilon_0, \varepsilon_1, \ldots, \varepsilon_m$ of $V'$-outcomes with $\varepsilon_0 = \alpha\bar{\alpha}$ and $\varepsilon_m = \beta\bar{\beta}$. Since $e$ holds in $\varepsilon_0$, (b) implies that the first flip changes some variable $y_i$, and (c) implies that the second flip changes variable $x_i$ to make $x_i \leftrightarrow y_i$ hold again. Hence $\varepsilon_2$ can be written as $\delta\bar{\delta}$. By (c) there is an improving flip in $C$ from outcome $\alpha$ changing variable $x_i$, that is, leading from $\alpha$ to $\delta$. Iterating this process shows that $L'(t)$ is an improving sequence from $\alpha$ to $\beta$.

Proof of (iii): Suppose that $C$ is inconsistent. Then there exists some outcome $\alpha$ and an improving sequence $s$ in $C$ from $\alpha$ to $\alpha$. By (i), $L(s)$ is an improving sequence from $\alpha\bar{\alpha}$ to $\alpha\bar{\alpha}$, proving that $C'$ is inconsistent.

Conversely, suppose that $C'$ is inconsistent, so there exists an improving sequence $t$ for $C'$ from some outcome to itself. By (a), any improving flip applied to an outcome in which $e$ does not hold increases (by one) the number of $i$ such that $x_i \leftrightarrow y_i$ holds. This implies that $e$ must hold in some outcome in $t$, because $t$ is not acyclic. Write this outcome as $\alpha\bar{\alpha}$. We can cyclically permute $t$ to form an improving sequence $t_2$ from $\alpha\bar{\alpha}$ to itself. Part (ii) then implies that $L'(t_2)$ is an improving flipping sequence for $C$ from $\alpha$ to itself, showing that $C$ is inconsistent. □

**Theorem 2** CP-DOMINANCE *is* PSPACE*-complete. This holds even if we restrict the CP-nets to being consistent.*

*Proof:* We use a reduction from PSPACE-hardness of the GCP-DOMINANCE problem when the GCP-nets are restricted to being consistent (Theorem 1). Let $C$ be a consistent, and hence locally consistent, GCP-net over $V$, and let $\alpha$ and $\beta$ be outcomes over $V$. Consider the CP-net $C'$ over variables $V'$ constructed above. Lemma 3(i) and (ii) imply that $\beta$ dominates $\alpha$ in $C$ if and only if $\beta\bar{\beta}$ dominates $\alpha\bar{\alpha}$ in $C'$. Moreover, $C'$ is consistent by Lemma 3(iii). Consequently, the hardness part of the assertion follows. □

Note that PSPACE-hardness obviously remains if we require input outcomes to be different, because the reduction for Theorem 1 uses a pair of different outcomes.

Notice the huge complexity gap with the problem of deciding whether there exists a non-dominated outcome, which is "only" NP-complete (Domshlak et al., 2003, 2006).

## 5. Consistency of GCP-Nets

In this section we show that the GCP-CONSISTENCY problem is PSPACE-complete, using results from Sections 3 and 4.

**Theorem 3**
GCP-CONSISTENCY *is* PSPACE-*complete. This holds even under the restriction to GCP-nets in conjunctive form.*

*Proof:* PSPACE-hardness is shown by reduction from ACTION-SET ACYCLICITY. We apply function $S$ from Section 3.2 followed by $M$ from Section 4.1. This maps instances of ACTION-SET ACYCLICITY to instances of GCP-CONSISTENCY in conjunctive form. By Lemma 1(iii) and Lemma 2 (ii), an instance of ACTION-SET ACYCLICITY is acyclic if and only if the corresponding instance of GCP-CONSISTENCY is consistent, proving the result. □

We now show that consistency testing remains PSPACE-complete for CP-nets (GCP-nets that are both locally consistent and locally complete).

**Theorem 4** CP-CONSISTENCY *is* PSPACE-*complete.*

*Proof:* We use a reduction from GCP-CONSISTENCY under the restriction that the GCP-net is in conjunctive form. Let $C$ be a GCP-net in conjunctive form. We define a CP-net $C'$ as follows. Because $C$ is in conjunctive form, local consistency can be decided in polynomial time, as it amounts to checking the consistency of a conjunction of conjunctions of literals. If $C$ is not locally consistent we set $C'$ to be a predetermined inconsistent but locally consistent CP-net, such as in the example in Section 2. Otherwise, $C$ is locally consistent and for $C'$ we take the CP-net we constructed in Section 4.2. The mapping from locally consistent GCP-nets to CP-nets, described in Section 4.2, preserves consistency (Lemma 3 (iii)). Since local inconsistency implies inconsistency (Proposition 2), we have that the GCP-net $C$ is consistent if and only if the CP-net $C'$ is consistent. Thus, PSPACE-hardness of the CP-CONSISTENCY problem follows from Theorem 3. □

## 6. Additional Problems Related to Dominance in GCP-Nets

Having proved our main results on consistency of and dominance in GCP-nets, we move on to additional questions concerning the dominance relation. Before we state them, we introduce more terminology.

Let $\alpha$ and $\beta$ be outcomes in a GCP-net $C$. We say that $\alpha$ and $\beta$ are *dominance-equivalent* in $C$, written $\alpha \approx_C \beta$, if $\alpha = \beta$, or $\alpha \prec_C \beta$ and $\beta \prec_C \alpha$. Next, $\alpha$ and $\beta$ are *dominance-incomparable* in $C$ if $\alpha \neq \beta$, $\alpha \not\prec_C \beta$ and $\beta \not\prec_C \alpha$. Finally, $\alpha$ *strictly dominates* $\beta$ if $\beta \prec_C \alpha$ and $\alpha \not\prec_C \beta$.

**Definition 8**
*We define the following decision problems:*
SELF-DOMINANCE: *given a GCP-net $C$ and an outcome $\alpha$, decide whether $\alpha \prec_C \alpha$, that is, whether $\alpha$ dominates itself in $C$.*
STRICT DOMINANCE: *given a GCP-net $C$ and outcomes $\alpha$ and $\beta$, decide whether $\alpha$ strictly dominates $\beta$ in $C$.*
DOMINANCE EQUIVALENCE: *given a GCP-net $C$ and outcomes $\alpha$ and $\beta$, decide whether $\alpha$ and $\beta$ are dominance-equivalent in $C$.*
DOMINANCE INCOMPARABILITY: *given a GCP-net $C$ and outcomes $\alpha$ and $\beta$, decide whether $\alpha$ and $\beta$ are dominance-incomparable in $C$.*

When establishing the complexity of these problems, we will use polynomial-time reductions from the problem GCP-DOMINANCE. Let $H$ be a GCP-net with the set of variables $V = \{x_1, \ldots, x_n\}$, and let $\beta$ be an outcome. We define a GCP-net $G = \Theta_1(H, \beta)$ with the set of variables $W = V \cup \{y\}$ by setting the conditions for flips on variables $x_i$, $i = 1, \ldots, n$, and $y$ as follows:

1. if $x_i \in \beta$:
   $p_G^+(x_i) = p_H^+(x_i) \vee \neg y$
   $p_G^-(x_i) = p_H^-(x_i) \wedge y$

2. if $\neg x_i \in \beta$:
   $p_G^+(x_i) = p_H^+(x_i) \wedge y$
   $p_G^-(x_i) = p_H^-(x_i) \vee \neg y$

3. $p_G^+(y) = \beta$

4. $p_G^-(y) = \neg \beta$.

The mapping $\Theta_1$ can be computed in polynomial time. Moreover, one can check that if $H$ is a locally consistent GCP-net, $\Theta_1(H, \beta)$ is also locally consistent. Finally, if $H$ is a CP-net, $\Theta_1(H, \beta)$ is a CP-net, as well.

For every $V$-outcome $\gamma$, we let $\gamma^+ = \gamma \wedge y$ and $\gamma^- = \gamma \wedge \neg y$. We note that every $W$-outcome is of the form $\gamma^+$ or $\gamma^-$. To explain the structure of the GCP-net $G$, we point out that there is an improving flip in $G$ from $\gamma^+$ into $\delta^+$ if and only if there is an improving flip in $H$ from $\gamma$ to $\delta$ (thus, $G$ restricted to outcomes of the form $\gamma^+$ forms a copy of the GCP-net $H$). Moreover, there is an improving flip in $G$ from $\gamma^-$ into $\delta^-$ if and only if $\delta$ agrees with $\beta$ on exactly one more variable $x_i$ than $\gamma$ does. Finally, an improving flip moves between outcomes of different type if and only if it transforms $\beta^-$ to $\beta^+$, or $\gamma^+$ to $\gamma^-$ for some $\gamma \neq \beta$.

We now formalize some useful properties of the GCP-net $G = \Theta_1(H, \beta)$. We use the notation introduced above.

**Lemma 4** *For every $V$-outcome $\gamma$, $\gamma^- \prec_G \beta^+$ and, if $\gamma \neq \beta$, $\gamma^+ \prec_G \beta^+$ (in other words, $\beta^+$ dominates every other $W$-outcome).*

*Proof:* Consider any $V$-outcome $\gamma \neq \beta$. Then $\gamma \wedge \neg y \prec_G \beta \wedge \neg y$ since, given $\neg y$, changing a literal to the form it has in $\beta$ is an improving flip. By the definition, we also have $\beta \wedge \neg y \prec_G \beta \wedge y$ and $\gamma \wedge y \prec_G \gamma \wedge \neg y$ (as $\gamma \neq \beta$). It follows that $\beta^- \prec_G \beta^+$ and $\gamma^+ \prec_G \gamma^- \prec_G \beta^+$. Thus, the assertion follows. □

**Lemma 5** *For arbitrary $V$-outcome $\alpha$ different from $\beta$, the following statements are equivalent:*

1. $\beta \prec_H \alpha$;

2. $\beta^+ \prec_G \alpha^+$;

3. $\beta^+ \approx_G \alpha^+$.

*Proof:* By Lemma 4, $\alpha^+ \prec_G \beta^+$. Thus, the conditions (2) and (3) are equivalent.

[(1)$\Rightarrow$(2)] Clearly (recall our discussion about the structure of $G$), if there is an improving flip from $\gamma$ to $\delta$ in $H$, then there is an improving flip from $\gamma^+$ to $\delta^+$ in $G$. Thus, if there is an improving sequence in $H$ from $\beta$ to $\alpha$, there is an improving sequence in $G$ from $\beta^+$ to $\alpha^+$.

[(2)$\Rightarrow$(1)] Let us assume $\beta^+ \prec_G \alpha^+$, and let us consider an improving sequence of minimum length from $\beta^+$ to $\alpha^+$. By the minimality, no internal element in such a sequence is $\beta^+$. Thus, no internal element equals $\beta^-$ either (as the only improving flip from $\beta^-$ leads to $\beta^+$). Since an improving flip from $\gamma^-$ to $\gamma^+$ requires that $\gamma = \beta$, all outcomes in the sequence are of the form $\gamma^+$. By dropping $y$ from each outcome in this sequence, we get an improving flipping sequence from $\alpha$ to $\beta$ in $H$. Thus, $\beta \prec_H \alpha$. □

**Lemma 6** *Let $H$ be consistent and let $\alpha$ and $\beta$ be different $V$-outcomes. Then, $\alpha^+ \prec_G \alpha^+$ if and only if $\beta \prec_H \alpha$.*

*Proof:* Suppose there exists an improving sequence from $\alpha^+$ to itself. There must be an outcome in the sequence of the form $\gamma \wedge \neg y$ (otherwise, dropping $y$ in every outcome yields an improving sequence from $\alpha$ to $\alpha$ in $H$, contradicting the consistency of $H$). To perform an improving flip from $\neg y$ to $y$ we need $\beta$ to hold, which implies that $\beta^+$ appears in the sequence. Thus, $\beta^+ \prec_G \alpha^+$. By Lemma 5, $\beta \prec_H \alpha$.

Conversely, let us assume that $\beta \prec_H \alpha$. Again by Lemma 5, $\beta^+ \prec_G \alpha^+$. By Lemma 4, $\alpha^+ \prec_G \beta^+$. Thus, $\alpha^+ \prec_G \alpha^+$. □

The next construction is similar. Let $H$ be a GCP-net on variables $V = \{x_1, \ldots, x_n\}$, and let $\alpha$ be an outcome. We define a GCP-net $F = \Theta_2(H, \alpha)$ as follows. As before, we set $W = V \cup \{y\}$ to be the set of variables of $F$. We define the conditions for flips on variables $x_i$, $i = 1, \ldots, n$, and $y$ as follows:

1. $p_G^+(x_i) = p_H^+(x_i) \wedge y$

2. $p_G^-(x_i) = p_H^-(x_i) \wedge y$

3. $p_G^+(y) = \neg \alpha$

4. $p_G^-(y) = \alpha$.

Informally, outcomes of the form $\gamma^+$ form in $F$ a copy of $H$. There are no improving flips between outcomes of the form $\gamma^-$. There is an improving flip from $\alpha^+$ to $\alpha^-$ and, for every $\gamma \neq \alpha$, from $\gamma^-$ to $\gamma^+$. In particular, if $F$ is consistent then $\Theta_2(H, \alpha)$ is consistent, The mapping $\Theta_2$ can be computed in polynomial time and we also have the following property.

**Lemma 7** *Let $\beta$ be a $V$-outcome different from $\alpha$. Then the following conditions are equivalent:*

1. $\beta \prec_H \alpha$

2. $\alpha^-$ *strictly dominates* $\beta^-$ *in $F$*

3. $\alpha^-$ *and* $\beta^-$ *are not dominance-incomparable in $F$.*

*Proof:* If there exists an improving sequence from $\beta^-$ to $\alpha^-$ then the first improving flip in the sequence changes $\beta^-$ to $\beta^+$. Moreover, there is an improving flip from $\gamma^+$ to $\gamma^-$ if and only if $\gamma = \alpha$. Thus, $\beta^- \prec_F \alpha^-$ if and only if $\beta \prec_H \alpha$. Since $\alpha^- \not\prec_F \beta^-$ all three conditions are equivalent. □

**Proposition 7** *The following problems are* PSPACE-*complete:* SELF-DOMINANCE, STRICT DOMINANCE, DOMINANCE EQUIVALENCE, *and* DOMINANCE INCOMPARABILITY.

*Proof:* For all four problems, membership is proven easily as for the problems in earlier sections.

For the PSPACE-hardness proofs, we use the problem CP-DOMINANCE in a version when we required that the input CP-net be consistent and the two input outcomes different. The problem is PSPACE-hard by Theorem 2.

Let $H$ be a consistent CP-net on a set $V$ of variables, and let $\alpha$ and $\beta$ be two different $V$-outcomes. By Lemma 5, $\beta \prec_H \alpha$ can be decided by deciding the problem DOMINANCE EQUIVALENCE for $\alpha^+$ and $\beta^+$ in the GCP-net $\Theta_1(H,\beta)$. Thus, the PSPACE-hardness of DOMINANCE EQUIVALENCE follows.

Next, the equivalence of Lemma 6, $\alpha^+ \prec_G \alpha^+ \Leftrightarrow \beta \prec_H \alpha$, which holds due to consistency of $H$, shows that the problem SELF-DOMINANCE is PSPACE-hard.

Finally, by Lemma 7, $\beta \prec_H \alpha$ can be decided either by deciding the problem STRICT DOMINANCE for outcomes $\alpha^-$ and $\beta^-$ in $\Theta_2(H,\alpha)$, or by deciding the complement of the problem DOMINANCE INCOMPARABILITY for $\alpha^-$ and $\beta^-$ in the GCP-net $\Theta_2(H,\alpha)$. It follows that STRICT DOMINANCE and DOMINANCE INCOMPARABILITY (the latter by the fact that coPSPACE=PSPACE) are PSPACE-complete.[8] □

**Corollary 1** *The problems* SELF-DOMINANCE *and* DOMINANCE EQUIVALENCE *are* PSPACE-*complete under the restriction to CP-nets. The problems* STRICT DOMINANCE *and* DOMINANCE INCOMPARABILITY *remain* PSPACE-*complete under the restriction to consistent CP-nets.*

*Proof:* Since in the proof of Proposition 7 we have that $H$ is a CP-net, the claim for the first two problems follows by our remarks that the mapping $\Theta_1$ preserves the property of being a CP-net.

For the last two problems, we observe that since $H$ in the proof of Proposition 7 is assumed to be consistent, $F = \Theta_2(H,\alpha)$ is consistent, too. Thus, it is also locally consistent and the mapping $F$ to $F'$ we used for the proof of Theorem 2 applies. In particular, $F'$ is a consistent CP-net and has the following properties (implied by Lemma 3):

1. $\alpha$ strictly dominates $\beta$ in $F$ if and only if $\alpha\overline{\alpha}$ strictly dominates $\beta\overline{\beta}$ in $F'$

2. $\alpha$ and $\beta$ are dominance-incomparable in $F$ if and only if $\alpha\overline{\alpha}$ and $\beta\overline{\beta}$ are dominance-incomparable in $F'$.

Since $F'$ is a consistent CP-net, the claim for the last two problems follows, too. □

---

8. For STRICT DOMINANCE, the result could have been also obtained as a simple corollary of Theorem 2, since in consistent GCP-nets dominance is equivalent to strict dominance.

## 7. Problems Concerning Optimality in GCP-Nets

The dominance relation $\prec_C$ of a GCP-net $C$ determines a certain order relation, which gives rise to several notions of optimality. We will introduce them and study the complexity of corresponding decision problems.

We first observe that the dominance equivalence relation is indeed an equivalence relation (reflexive, symmetric and transitive). Thus, it partitions the set of all outcomes into non-empty equivalence classes, which we call *dominance* classes. We denote the dominance class of an outcome $\alpha$ in a GCP-net $C$ by $[\alpha]_C$.

The relation $\prec_C$ induces on the set of dominance classes a *strict order* relation (a relation that is irreflexive and transitive). Namely, we define $[\alpha]_C \prec_C^{dc} [\beta]_C$ if $[\alpha]_C \neq [\beta]_C$ (equivalently, $\alpha \not\approx_C \beta$) and $\alpha \prec_C \beta$. One can check that the definition of the relation $\prec_C^{dc}$ on dominance classes is independent of the choice of representatives of the classes.

**Definition 9 (Non-dominated class, optimality in GCP-nets)** *Let $C$ be a GCP-net. A dominance class $[\alpha]_C$ is* non-dominated *if it is maximal in the strict order $\prec_C^{dc}$ (there is no dominance class $[\beta]_C$ such that $[\alpha]_C \prec_C^{dc} [\beta]_C$). A dominance class is* dominating *if for every dominance class $[\beta]_C$, $[\alpha]_C = [\beta]_C$ or $[\beta]_C \prec_C^{dc} [\alpha]_C$.*

*An outcome $\alpha$ is* weakly non-dominated *if it belongs to a non-dominated class. If $\alpha$ is weakly non-dominated and is the only element in its dominance class, then $\alpha$ is* non-dominated.

*An outcome $\alpha$ is* dominating *if it belongs to a dominating class. An outcome $\alpha$ is* strongly dominating *if it is dominating and non-dominated.*

Outcomes that are weakly non-dominated, non-dominated, dominating and strongly dominating capture some notions of optimality. In the context of CP-nets, weakly non-dominated and non-dominated outcomes were proposed and studied before (Brafman & Dimopoulos, 2004). They were referred to as weakly and strongly optimal there. Similar notions of optimality were also studied earlier for the problem of defining winners in partial tournaments (Brandt, Fischer, & Harrenstein, 2007). We will study here the complexity of problems to decide whether a given outcome is optimal and whether optimal outcomes exist.

First, we note the following general properties (simple consequences of properties of finite strict orders).

**Lemma 8** *Let $C$ be a GCP-net.*

1. *There exist non-dominated classes and so, weakly non-dominated outcomes.*

2. *Dominating outcomes and nondominated outcomes are weakly non-dominated.*

3. *A strongly dominating outcome is dominating and non-dominated.*

4. *The following conditions are equivalent:*

    *(a) $C$ has a unique non-dominated class;*
    
    *(b) $C$ has a dominating outcome;*
    
    *(c) weakly non-dominated and dominating outcomes in $C$ coincide.*

For consistent GCP-nets only two different notions of optimality remain.

**Lemma 9** *Let C be a consistent GCP-net. Then:*

1. *Each dominance class is a singleton, $\prec_C$ is a strict order, and $\prec_C$ and $\prec_C^{dc}$ coincide (modulo the one-to-one and onto correspondence $\alpha \mapsto [\alpha]_C$)*

2. *If $\alpha$ is a weakly non-dominated outcome, $\alpha$ is non-dominated (weakly non-dominated and non-dominated outcomes coincide)*

3. *If $\alpha$ is a dominating outcome, $\alpha$ is strongly dominating (strongly dominating and dominating outcomes coincide).*

4. *Finally, $\alpha$ is a unique (weakly) non-dominated outcome if and only if $\alpha$ is strongly dominating.*

Next, we observe that all concepts of optimality we introduced are different. To this end, we will show GCP-nets with a single non-dominated class that is a singleton, with multiple non-dominated classes, each being a singleton, with a single non-dominated class that is not a singleton, and with multiple non-dominated classes, each containing more than one element. We will also show a GCP-net with two non-dominated classes, one of them a singleton and the other one consisting of several outcomes.

**Example 2** *Consider the following GCP-net C with two binary variables a and b*

$: a > \bar{a}$
$: b > \bar{b}$

*This GCP-net determines a strict preorder on the dominance classes, in which $\{ab\}$ is the only maximal class (in fact, all dominance classes are singletons). Thus, ab is both non-dominated and dominating and so, it is strongly dominating.*

**Example 3** *Consider the following GCP-net C with two binary variables a and b*

$b : a > \bar{a}$
$\bar{b} : \bar{a} > a$
$a : b > \bar{b}$
$\bar{a} : \bar{b} > b$

*This GCP-net determines a strict preorder, in which $\{ab\}$ and $\{\bar{a}\bar{b}\}$ are two different non-dominated classes. Thus, ab and $\bar{a}\bar{b}$ are non-dominated and there is no dominating outcome.*

**Example 4** *Consider a GCP-net with variables a,b and c, defined as follows:*

$a : b > \bar{b}$
$\bar{a} : \bar{b} > b$
$\bar{b} : a > \bar{a}$
$b : \bar{a} > a$
$ab : c > \bar{c}$

*There are two dominance classes: $S_c = \{abc, a\bar{b}c, \bar{a}bc, \bar{a}\bar{b}c\}$ and $S_{\bar{c}} = \{ab\bar{c}, a\bar{b}\bar{c}, \bar{a}b\bar{c}, \bar{a}\bar{b}\bar{c}\}$. Every outcome in $S_c$ strictly dominates every outcome in $S_{\bar{c}}$, therefore, $S_c$ is the unique non-dominated class and every outcome in $S_c$ is dominating. Because $S_c$ is not a singleton, there are no non-dominated outcomes (and so, no strongly dominating outcome, either).*

**Example 5** *Let us remove from the GCP-net of Example 4 the preference statement $ab : c > \bar{c}$. Then $S_c$ and $S_{\bar{c}}$ are still the two dominance classes, but now every outcome is $S_c$ is incomparable with any outcome in $S_{\bar{c}}$. Thus, $S_c$ and $S_{\bar{c}}$ are both non-dominated. Since there are two non-dominated classes, there is no dominating outcome. Since each class has more than one element, there are no non-dominated outcomes. All outcomes are weakly non-dominated, though.*

**Example 6** *Let us modify the GCP-net of Example 4 by changing the preference statement $\bar{b} : a > \bar{a}$ into $\bar{b}c : a > \bar{a}$. The dominance relation $\prec$ of this GCP-net satisfies the following properties: (i) the four outcomes in $S_c$ dominate each other; (ii) $\bar{a}\bar{b}\bar{c} \succ \bar{a}b\bar{c} \succ ab\bar{c} \succ a\bar{b}\bar{c}$; (iii) any outcome in $S_c$ dominates $ab\bar{c}$ (and, a fortiori, $a\bar{b}\bar{c}$). One can check that there are five dominance classes: $S_c$, $\{ab\bar{c}\}$, $\{\bar{a}b\bar{c}\}$, $\{a\bar{b}\bar{c}\}$ and $\{\bar{a}\bar{b}\bar{c}\}$. Two of them are non-dominated: $S_c$ and $\{\bar{a}\bar{b}\bar{c}\}$. Since there are two non-dominated classes, there is no dominating outcome. On the other hand, $\{\bar{a}\bar{b}\bar{c}\}$ is a non-dominated outcome (a unique one).*

We will consider the following decision problems corresponding to the notions of optimality we introduced.

**Definition 10**
*For a given GCP-net C:*
WEAKLY NON-DOMINATED OUTCOME: *given an outcome $\alpha$, decide whether $\alpha$ is weakly non-dominated in C*
NON-DOMINATED OUTCOME: *given an outcome $\alpha$, decide whether $\alpha$ is non-dominated in C*
DOMINATING OUTCOME: *given an outcome $\alpha$, decide whether $\alpha$ is dominating in C*
STRONGLY DOMINATING OUTCOME: *given an outcome $\alpha$, decide whether $\alpha$ is strongly dominating in C*
EXISTENCE OF A NON-DOMINATED OUTCOME: *decide whether C has a non-dominated outcome*
EXISTENCE OF A DOMINATING OUTCOME: *decide whether C has a dominating outcome*
EXISTENCE OF A STRONGLY DOMINATING OUTCOME: *decide whether C has a strongly dominating outcome.*

In some of the hardness proofs, we will again use the reductions $\Theta_1$ and $\Theta_2$, described in the previous section. We note the following additional useful properties of the GCP-net $G = \Theta_1(H, \beta)$.

**Lemma 10** *For arbitrary V-outcome $\alpha$ different from $\beta$, the following statements are equivalent:*

1. $\beta^+ \prec_G \alpha^+$

2. $\alpha^+$ *is weakly non-dominated in G*

3. $\alpha^+$ *is a dominating outcome in G.*

*Proof:* Since $\beta^+$ is dominating in $G$ (Lemma 4), weakly non-dominated outcomes and dominating outcomes coincide (Lemma 8). It follows that the conditions (1)-(3) are equivalent to each other. □

**Proposition 8** *The following problems are* PSPACE-*complete:* WEAKLY NON-DOMINATED OUTCOME *and* DOMINATING OUTCOME. *The result holds also for the problems restricted to CP-nets.*

*Proof:* The membership is easy to prove by techniques similar to those we used earlier.

For the PSPACE-hardness proofs, we use reductions from CP-DOMINANCE for consistent CP-nets (in the version where the two input outcomes are different). Let $H$ be a CP-net, and $\alpha$ and $\beta$ two different $V$-outcomes. By Lemmas 5 and 10, $\beta \prec_H \alpha$ can be decided by deciding either of the problems WEAKLY NON-DOMINATED OUTCOME and DOMINATING OUTCOME for the GCP-net $G = \Theta_1(H, \beta)$ and the outcome $\alpha^+$. We observed earlier, that if $H$ is a CP-net, then so is $G = \Theta_1(H, \beta)$. Thus, the second part of the assertion follows. □

Next, we will consider the problem STRONGLY DOMINATING OUTCOME. We will exploit the reduction $F = \Theta_2(H, \alpha)$, which we discussed in the previous section. We observe the following property of $F$.

**Lemma 11** *Let $H$ be a GCP-net and $F = \Theta_2(H, \alpha)$. Then $\alpha^-$ is strongly dominating in $F$ if and only if $\alpha$ is dominating in $H$.*

*Proof:* Let us assume that $\alpha$ is dominating in $H$. From the definition of $F$, it follows that for every $V$-outcome $\gamma \neq \alpha$, $\gamma^+ \prec_F \alpha^+$ and $\gamma^- \prec_F \gamma^+$. Since $\alpha^+ \prec_F \alpha^-$, $\alpha^-$ is dominating in $F$. Since there is no improving flip leading out of $\alpha^-$, $\alpha^-$ is strongly dominating.

Conversely, let us assume that $\alpha^-$ is strongly dominating in $F$ and let $\gamma$ be a $V$-outcome different from $\alpha$. Let us consider an improving sequence from $\gamma^+$ to $\alpha^-$. All outcomes in the sequence other than the last one, $\alpha^-$, are of the form $\delta^+$. Moreover, the outcome directly preceding $\alpha^-$ is $\alpha^+$. Dropping $y$ from every outcome in the segment of the sequence between $\gamma^+$ and $\alpha^+$ yields an improving sequence from $\gamma$ to $\alpha$ in $H$. □

We now have the following consequence of this result.

**Proposition 9** *The problem* STRONGLY DOMINATING OUTCOME *is* PSPACE-*complete, even if restricted to CP-nets.*

*Proof:* Let $H$ be a CP-net (over the set $V$ of variables) and $\alpha$ an outcome. By Lemma 11, the problem DOMINATING OUTCOME can be decided by deciding the problem STRONGLY DOMINATING OUTCOME for $F = \Theta_2(H, \alpha)$ and $\alpha^-$. Thus, the PSPACE-hardness of STRONGLY DOMINATING OUTCOME follows by Proposition 8. The membership in PSPACE is, as in other cases, standard and is omitted.

Since $H$ is a CP-net, it is locally consistent and so, $F$ is locally consistent, too. As in the proof of Corollary 1 we use the mapping from GCP-net $F$ to CP-net $F'$ defined in Section 4.2. By Lemma 3, $\alpha$ is a strongly dominating outcome in $F$ if and only if $\alpha\bar{\alpha}$ dominates every outcome of the form $\gamma\bar{\gamma}$, which is if and only if $\alpha\bar{\alpha}$ is a strongly dominating outcome in $F'$, since any $F'$-outcome is dominated by an outcome of the form $\gamma\bar{\gamma}$ (using the rules $q^+(x_i) = y_i$ and $q^-(x_i) = \neg y_i$). Therefore

STRONGLY DOMINATING OUTCOME for $F$ and $\alpha$ can be decided by deciding STRONGLY DOMINATING OUTCOME for $F'$ and $\alpha\overline{\alpha}$. Thus, the second part of the claim follows. □

The problem NON-DOMINATED OUTCOME is easier. It is known to be in P for CP-nets (Brafman & Dimopoulos, 2004). The result extends to GCP-nets. Indeed, if $H$ is a GCP-net and $\alpha$ an outcome, $\alpha$ is non-dominated if and only if there is no improving flip that applies to $\alpha$. The latter holds if and only if for every variable $x$ in $H$, if $x$ (respectively, $\neg x$) holds in $\alpha$, then $p^-(x)$ (respectively, $p^+(x)$) does not hold in $\alpha$. Since the conditions can be checked in polynomial the claim holds and we have the following result.

**Proposition 10** *The problem* NON-DOMINATED OUTCOME *for GCP-nets is in* P.

Next, we will consider the problems concerning the existence of optimal outcomes. Let $H$ be a GCP-net on the set of variables $V = \{x_1, \ldots, x_n\}$, and let $\alpha$ and $\beta$ be two different $V$-outcomes. For every $i = 1, 2, \ldots, n$, we define formulas $\alpha_i$ as follows. If $x_i \in \alpha$, then $\alpha_i$ is the conjunction of all literals in $\alpha$, except that instead of $x_i$ we take $\neg x_i$. Similarly, if $\neg x_i \in \alpha$, then $\alpha_i$ is the conjunction of all literals in $\alpha$, except that instead of $\neg x_i$ we take $x_i$. Thus, $\alpha_i$ is the outcome that results in $\alpha$ when the literal in corresponding to $x_i$ is flipped into its dual.

We now define a GCP-net $E = \Theta_3(H, \alpha, \beta)$ by taking $W = V \cup \{y\}$ as the set of variables of $E$ and by defining the flipping conditions as follows:

1. $p_E^+(x_i) = (p_H^+(x_i) \wedge y) \vee (\neg y \wedge \neg \alpha \wedge \neg \alpha_i)$
   $p_E^-(x_i) = p_H^-(x_i) \wedge y$

2. $p_E^+(y) = \beta$

3. $p_E^-(y) = \neg\beta$.

The GCP-net $\Theta_3(H, \alpha, \beta)$ has the following properties. The outcomes of the form $\gamma^+ (= \gamma \wedge y)$ form a copy of $H$. There is no improving flip for the outcome $\alpha^- (= \alpha \wedge \neg y)$. Next, there is no improving flip into $\alpha^-$ from an outcome of the form $\gamma^-$. To see this, let us assume that such a flip exists and concerns a variable, say, $x_i$. It follows that $\gamma = \alpha_i$. By the definition of flipping conditions, an improving flip for $\gamma^-$ that involves $x_i$ is impossible, a contradiction. Thus, the only improving flip that leads to $\alpha^-$ originates in $\alpha^+$.

We also have that for every outcome $\gamma$ other than $\alpha$ and $\beta$, $\gamma^- \prec_E \beta^-$. It follows from the fact that for every outcome $\gamma$ other than $\alpha$ and $\beta$, $\gamma^-$ has an improving flip. Indeed, for each such $\gamma$ there is a variable $x_i$ such that (i) $x_i$ is false in $\gamma$, and (ii) flipping the literal of $x_i$ to its dual does not lead to $\alpha$ (that is, $\gamma$ is not $\alpha_i$). (For even if $\gamma = \alpha_i$ for some $i$, then, because $\gamma, \alpha \neq \beta$, there exists $i' \neq i$ such that $\gamma$ and $\beta$ differ on $x_{i'}$, so that $x_{i'}$ satisfies (i) and (ii).) Thus, a flip on that variable is improving. As all improving flips between outcomes containing $\neg y$ result in one more variable $x_i$ assigned to true, thus having the same status as it has in $\beta$, $\gamma^- \prec_E \beta^-$ follows.

Finally, we have $\beta^- \prec_E \beta^+$ and, for every outcome $\gamma$ other than $\beta$, $\gamma^+ \prec_E \gamma^-$. This leads to the following property of $E = \Theta_3(H, \alpha, \beta)$.

**Lemma 12** *Let $H$ be a GCP-net and let $\alpha$ and $\beta$ be two different outcomes. Then $\beta \prec_H \alpha$ if and only if $\Theta_3(H, \alpha, \beta)$ has a (strongly) dominating outcome.*

*Proof:* (Only if) Based on our earlier remarks, $\alpha^+ \prec_E \alpha^-$. Moreover, since $\beta \prec_H \alpha$, we have $\beta^+ \prec_E \alpha^+$. In addition, for every $\gamma$ different from $\alpha$ and $\beta$, $\gamma^+ \prec_E \gamma^- \prec_E \beta^- \prec_E \beta^+$. Thus, $\alpha^-$ is both dominating and strongly dominating (the latter follows from the fact that no improving flips lead out of $\alpha^-$).

(If) Let us assume that $\alpha^-$ is dominating (and so, the argument applies also when $\alpha^-$ is strongly dominating). Then there is an improving sequence from $\beta^+$ to $\alpha^-$. Let us consider a shortest such sequence. Clearly, $\alpha^+$ is the outcome just before $\alpha^-$ in that sequence (as we pointed out, no improving flip from an outcome of the form $\gamma^-$ to $\alpha^-$ is possible). Moreover, by the definition of $\Theta_3(H, \alpha, \beta)$ and the fact that we are considering a shortest sequence from $\beta^+$ to $\alpha^-$, every outcome in the sequence between $\beta^+$ and $\alpha^+$ is of the form $\gamma^+$. By dropping $y$ from each of these outcomes, we get an improving sequence from $\beta$ to $\alpha$. □

**Proposition 11** *The problem* EXISTENCE OF DOMINATING OUTCOME *and the problem* EXISTENCE OF STRONGLY DOMINATING OUTCOME *are* PSPACE-*complete, even if restricted to CP-nets.*

*Proof:* We show the hardness part only, as the membership part is straightforward. To prove hardness we notice that by Lemma 12, given a consistent CP-net $H$ and two outcomes $\alpha$ and $\beta$, $\beta \prec_H \alpha$ can be decided by deciding either of the problems EXISTENCE OF DOMINATING OUTCOME and EXISTENCE OF STRONGLY DOMINATING OUTCOME for $\Theta_3(H, \alpha, \beta)$. To prove the second part of the assertion, we note that if $H$ is consistent, $E = \Theta_3(H, \alpha, \beta)$ is consistent, too and so, the mapping from locally consistent GCP nets to CP-nets applies. Let us denote the result of applying the mapping to $E$ by $E'$. Then, using the same argument as in the proof of Proposition 9, $E$ has a (strongly) dominating outcome if and only if $E'$ has a strongly dominating outcome. Thus, one can decide whether $\beta \prec_H \alpha$ in a consistent CP-net $H$ by deciding either of the problems EXISTENCE OF DOMINATING OUTCOME and EXISTENCE OF STRONGLY DOMINATING OUTCOME for $E'$. □

We also note that the problem EXISTENCE OF NON-DOMINATED OUTCOME is easier (under standard complexity theory assumptions).

**Proposition 12** *The problem* EXISTENCE OF NON-DOMINATED OUTCOME *is* NP-*complete.*

*Proof:* We note that in the case of GCP-nets in conjunctive form the problem is known to be NP-hard (Domshlak et al., 2003, 2006). Thus, the problem is NP-hard for GCP-nets. The membership in the class NP follows from Proposition 10. □

If we restrict to consistent GCP-nets, the situation simplifies. First, we recall (Lemma 9) that if a GCP-net is consistent then weakly non-dominated and non-dominated outcomes coincide, and the same is true for dominating and strongly dominating outcomes. Moreover, for consistent GCP-nets, non-dominated outcomes exist (and so, the corresponding decision problem is trivially in P). Thus, for consistent GCP-nets we will only consider problems DOMINATING OUTCOME and EXISTENCE OF DOMINATING OUTCOME.

**Proposition 13** *The problems* DOMINATING OUTCOME *and* EXISTENCE OF DOMINATING OUTCOME *restricted to consistent GCP-nets are in* coNP.

*Proof:* Using Lemmas 8 and 9, $\alpha$ is not a dominating outcome if and only if there exists an outcome $\beta \neq \alpha$ which is non-dominated. Similarly, there is no dominating outcome in a consistent GCP-net if and only if there are at least two non-dominated outcomes. Thus, guessing non-deterministically an outcome $\beta \neq \alpha$, and verifying that $\beta$ is non-dominated, is a non-deterministic polynomial-time algorithm deciding the complement of the problem DOMINATING OUTCOME. The argument for the other problem is similar. □

We do not know if the bounds in Proposition 13 are tight, that is, whether these two problems are coNP-complete. We conjecture they are.

## 8. Concluding Remarks

We have shown that dominance and consistency testing in CP-nets are both PSPACE-complete. Also several related problems related to dominance and optimality in CP-nets are PSPACE-complete, too.

The repeated use of reductions from planning problems confirms the importance of the structural similarity between STRIPS planning and reasoning with CP-nets. This suggests that the well-developed field of planning algorithms for STRIPS representations, especially for unary operators (Brafman & Domshlak, 2003), could be useful for implementing algorithms for dominance and consistency in CP-nets.

Our theorems extend to CP-nets with non-binary domains, and to extensions and variations of CP-nets, such as TCP-nets (Brafman & Domshlak, 2002; Brafman, Domshlak, & Shimony, 2006) that allow for explicit priority of some variables over others, and the more general language for conditional preferences (Wilson, 2004a, 2004b), where the conditional preference rules are written in conjunctive form.

The complexity result for dominance is also relevant for the following constrained optimisation problem: given a CP-net and a constraint satisfaction problem (CSP), find an optimal solution (a solution of the CSP which is not dominated by any other solution of the CSP). This is computationally complex, intuitively because a complete algorithm involves many dominance checks when the definition of dominance under constraints allows for dominance paths to go through outcomes violating the constraints (Boutilier, Brafman, Domshlak, Hoos, & Poole, 2004b).[9] The problem of checking whether a given solution of a CSP is non-dominated can be seen to be PSPACE-complete by a reduction from CP-dominance that uses a CSP that has exactly two solutions.

Our results reinforce the need for work on finding special classes of problems where dominance and consistency can be tested efficiently (Domshlak & Brafman, 2002; Boutilier et al., 2004a), and for incomplete methods for checking consistency and constrained optimisation (Wilson, 2004a, 2006).

Several open problems remain. We do not know the complexity of deciding whether the preference relation induced by a CP-net is complete. We do not know whether dominance and consistency testing remain PSPACE-complete when the number of parents in the dependency graph is bounded by a constant. We also do not know whether these two problems remain PSPACE-complete for CP-nets in conjunctive form (the reduction used to prove Theorems 2 and 4 yields CP-nets that are not in conjunctive form). Two additional open problems are listed at the end of Section 7.

---

9. With another possible definition, where going through outcomes violating the constraints is not allowed (Prestwich, Rossi, Venable, & Walsh, 2005), dominance testing is not needed to check whether a given solution is non-dominated.

**Acknowledgments**

Jérôme Lang's new address is: LAMSADE, Université Paris-Dauphine, 75775 Paris Cedex 16, France. The authors are grateful to the reviewers for their excellent comments, and to Pierre Marquis for helpful discussions. This work was supported in part by the NSF under Grants ITR-0325063, IIS-0097278 and KSEF-1036-RDE-008, by the ANR Project ANR–05–BLAN–0384 "Preference Handling and Aggregation in Combinatorial Domains", by Science Foundation Ireland under Grants No. 00/PI.1/C075 and 05/IN/I886, and by Enterprise Ireland Ulysses travel grant FR/2006/36.